%% file: DGML.tex
\documentclass{article}

\usepackage{times}
\usepackage{graphicx} 
\usepackage{subcaption}
\usepackage{natbib}

\usepackage{algorithm}
\usepackage{algorithmic}

\usepackage{hyperref}

\usepackage[accepted]{icml2017} 

\usepackage{amsmath}
\usepackage{amsfonts}
\usepackage{tikz}
\usepackage{bm}
\usepackage{standalone}
\usetikzlibrary{bayesnet}
\usetikzlibrary{arrows}
\usepackage{float}

\icmltitlerunning{Bayesian Semisupervised Learning}

\begin{document} 

\twocolumn[
\icmltitle{Bayesian Semisupervised Learning with Deep Generative Models}



\icmlsetsymbol{equal}{*}

\begin{icmlauthorlist}
\icmlauthor{Jonathan Gordon}{}
\icmlauthor{Jos\'e Miguel Hern\'andez-Lobato}{}
\end{icmlauthorlist}



\icmlcorrespondingauthor{Jonathan Gordon}{jg801@cam.ac.uk}

\icmlkeywords{Probabilistic deep learning, Deep generative models, active learning, semisupervised learning}

\vskip 0.3in
]




\begin{abstract} 
Neural network based generative models with discriminative components are a powerful approach for semi-supervised learning. However, these techniques a) cannot account for model uncertainty in the estimation of the model's discriminative component and b) lack flexibility to capture complex stochastic patterns in the label generation process. To avoid these problems, we first propose to use a discriminative component with stochastic inputs for increased noise flexibility. We show how an efficient Gibbs sampling procedure can marginalize the stochastic inputs when inferring missing labels in this model. Following this, we extend the discriminative component to be fully Bayesian and produce estimates of uncertainty in its parameter values. This opens the door for semi-supervised Bayesian active learning.
\end{abstract}

\section{Introduction}

\begin{figure*}[ht]
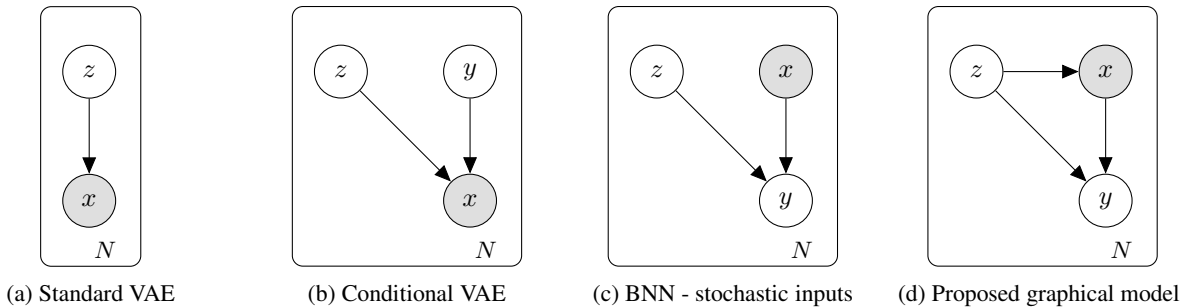

\vskip 0.2in
\begin{center}
	\centering
    \begin{subfigure}{.24\textwidth}
		\centering
  		\includestandalone{vae}
    	\subcaption{Standard VAE}
		\label{fig:vae_graph}
	\end{subfigure} %
	\begin{subfigure}{.24\textwidth}
    	\centering
  		\includestandalone{cvae}
    	\subcaption{Conditional VAE}
		\label{fig:cvae_graph}
    \end{subfigure} %
	\begin{subfigure}{.24\textwidth}
    	\centering
  		\includestandalone{stochastic_bnn}
    	\subcaption{BNN - stochastic inputs}
		\label{fig:bnn_graph}
	\end{subfigure} 
    \begin{subfigure}{.24\textwidth}
    \centering
    \scalebox{1.0}{
	\centering
	\tikz{
		\node[obs] (x) {$x$};%
		\node[latent,left=of x] (z) {$z$}; %
		\node[latent,below=of x] (y) {$y$}; %
		\plate [inner sep=.3cm,xshift=.02cm,yshift=.2cm] {plate1} {(x) (z) (y)} {$N$}; %
		\edge {z, x} {y} ; %
		\edge {z} {x}
		}}
    \subcaption{Proposed graphical model}
    \label{fig:our_graph}
    \end{subfigure}
	\caption{Graphical model depiction for VAE and BNN based models and proposed approach.}
	\label{fig:graphical_models}
\end{center}
\end{figure*}

Deep generative models (DGMs) can model complex high dimensional data via the use of latent variables. Recently, advances in variational training procedures such as stochastic backpropagation \cite{rezende2014stochastic} and the reparametrization trick \cite{kingma2013auto} have made training these models feasible and efficient. DGMs are particularly powerful when neural networks are used to parameterize generative distributions and inference networks, leading to the Variational Autoencoder (VAE, Figure \ref{fig:vae_graph})  \cite{kingma2013auto,rezende2014stochastic}.

The ability to efficiently train such models has led to a plethora of interesting extensions, increasing the flexibility of posterior approximations, expressiveness, and capabilities of the models \cite{burda2015importance,tomczak2016improving,maaloe2016auxiliary,rezende2015variational}. An important extension to standard VAEs is for semi-supervised learning \cite{kingma2014semi,maaloe2016auxiliary}, which incorporates labels into the generative model of the inputs (Figure \ref{fig:cvae_graph}), extending the VAE to semi-supervised learning tasks. In this settting, the labels are treated as latent variables that influence the generative process for the inputs and a recognition network is then used as a discriminative component to infer missing labels.

One major drawback of these previous approaches for semi-supervised learning with VAEs is that after training, the generative model is discarded and the recognition network is the only element used as a discriminative component \cite{kingma2014semi, maaloe2016auxiliary}. This is unsatisfactory from a modeling perspective as the recognition network is just a tool for performing approximate inference, and cannot be used to quantify model uncertainty.

Closely related to deep generative models are Bayesian neural networks (BNNs) \cite{neal2012bayesian}. BNNs extend standard neural networks and explicitly model the uncertainty in the learned weights increasing robustness and opening the door to tasks requiring uncertainty such as active learning \citep{gal2017deep,Hernandez-Lobato2015} and Bayesian optimization \cite{snoek2015scalable}. \citet{blundell2015weight} extend ideas from stochastic variational inference \cite{kingma2013auto,ranganath2014black} to an efficient inference procedure for BNNs. Further, \citet{depeweg2016learning} show how learning can be extended to general $\alpha$-divergence minimization \cite{hernandez2016black,moerland2017learning} and provide empirical evidence of the benefit of introducing stochastic inputs (Figure \ref{fig:bnn_graph}). However, in these works the stochastic inputs are of low (typically one) dimension.

We extend these ideas and develop a deep generative model with a discriminative component given by a BNN with stochastic inputs to accommodate semi-supervised learning. Our motivation is that after training, this BNN can be used to infer missing labels, rather than using the inference networks. This allows us to fully quantify any modeling uncertainty in the predictions of our discriminative component. We introduce two recognition networks and demonstrate how they can be used for training and prediction, as well as for posterior inference of high-dimensional stochastic inputs at prediction time. Our goal is to use the proposed method for semi-supervised Bayesian active learning.

\section{Related Work}\label{sec:related_work}

DGMs have recently shown to be very effective in semi-supervised learning tasks \citep{kingma2014semi,maaloe2016auxiliary}, achieving state-of-the-art performance on a number of benchmarks. Our model is most similar to the work detailed by \citet{kingma2014semi}. However, since our 
discriminative component is a Bayesian neural network with random inputs, we use a slightly different inference network architecture.

Similarly, Bayesian deep learning has recently been shown to be highly effective in active learning regimes \citep{Hernandez-Lobato2015,gal2017deep}. In contrast to these works, the proposed model can perform semi-supervised and active learning simultaneously, which may lead to significant improvements. Another difference is that while \citet{gal2017deep} use dropout as a proxy for Bayesian inference \citep{gal2015dropout} and \citet{Hernandez-Lobato2015} use a technique called probabilistic backpropagation, we propose leveraging variational inference to explicitly model the weight uncertainty \cite{blundell2015weight}.

The proposed model builds on ideas from both DGMs and Bayesian deep networks to suggest a principled method for simultaneous semi-supervised and active learning.

\section{Deep Generative Model of Labels}\label{sec:model}

We propose extending the model developed by \citet{depeweg2016learning} (as in Fig. \ref{fig:our_graph}) and including an inference network for $z$, allowing scalability of the latent dimension. Further, we propose inference procedures to allow this model to be used for semi-supervised learning, similarly to \citet{kingma2014semi,maaloe2015improving}.

There are a few motivations for our approach: (i) it builds on the idea of VAEs, but does so in a manner that results in an explicit probabilistic model for the labels, (ii) it extends BNNs with stochastic inputs to include inference networks for the stochastic inputs, allowing generalizing these to high dimensional variables, and (iii) it naturally accommodates semi-supervised and active learning with the generative model. The generative model can be described as:
\allowdisplaybreaks
\begin{align}\label{eqn:model}
\allowdisplaybreaks
p(z) & = \mathcal{N}\left(z; \bm{0}, \bm{I}\right)\,,\\
p_{\theta}(x|z) & = \mathcal{N}\left(x; \mu_x, \nu_x \right)\,,\\
p_{\theta}(y|z,x) & = \text{Cat}\left(y; \pi_y\right)\,,
\end{align}
where we parameterize the distributions of $x,y$ with deep neural networks:
\begin{align}
\mu_x  & =  NN_x(z, \theta)\,,& \log\nu_x & = NN_x(z, \theta)\,,\\ \pi_y & = NN_y(z, x, \theta)\,,
\end{align}
where $NN_{x}$ and $NN_y$ have weights $\mathcal{W}_{x}$ and $\mathcal{W}_y$ respectively, and $\theta = \{\mathcal{W}_x, \mathcal{W}_y\}$.

\subsection{Variational Training of the Model}

We propose a variational approach for training the model with both labeled and unlabeled data. In this section we are interested in point estimates of $\theta$ and Bayesian inference for $z$. We first derive the variational lower bound. In the semi-supervised setting, there are two lower bounds (ELBOs), for the labeled and unlabeled case. 

\subsubsection{Labeled Data ELBO}

\noindent Following recent advances in variational inference \cite{kingma2013auto,rezende2014stochastic}, we introduce inference networks $q_{\phi}(z|x,y)$ to approximate the intractable posterior distribution. Taking expectations w.r.t. $q$ of the log likelihoods we have:

\vspace{-0.35cm}
{\small
\begin{align}\label{eqn:elbo_l}
\log p_{\theta}(x,y)  & \geq  \mathbb{E}_{q_{\phi(z|x,y)}}  \left[\log p_{\theta}(x|z) + \log p_{\theta}(y|x,z) \right] - \nonumber\\ 
& \quad \,\,\text{D}_\text{KL}\left(q_{\phi}(z | x,y) || p(z)  \right) = \mathcal{L}^l (x,y;\theta, \phi) \,,
\end{align}
}where $q_{\phi}(z | x,y)$ is a recognition network parameterized by $\phi$. The lower-bound contains a term for the likelihood associated with the pair of variables $x$ and $y$, and a regularization term for the inference network. We can approximate expectations w.r.t. $q_{\phi}(z|x,y)$ via Monte Carlo estimation:
\begin{align}
\mathcal{L}^l (x,y; \theta, \phi) & \approx \frac{1}{L}\sum\limits_{l=1}^L \left[\log p_{\theta}(x,y|z^l)-\right.\nonumber\\ 
&\quad\,\,\left.\log q_{\phi}(z^l|x) + \log p(z^l)\right]\,,
\end{align}
with $z^l\sim q_{\phi}(z|x,y)$, and for Gaussian recognition networks the KL term can be evaluated analytically. 

\subsubsection{Unlabeled Data ELBO}

We can follow a similar approach to derive the lower bound for the unlabeled case. In this setting, we have:
\begin{align}\label{eqn:elbo_u2}
\log p_{\theta}(x) & \geq \mathbb{E}_{q_{\phi}(y|x)}\left[ \mathcal{L}^l (x, y; \theta, \phi) \right] + \mathcal{H}\left[q_{\phi}(y|x)\right] =\nonumber\\
& \quad \,\, \mathcal{L}^u(x; \theta, \phi)\,,
\end{align}
where we have used the decomposition $q_{\phi}(z,y|x) = q_{\phi}(y|x)q_{\phi}(z|x,y)$, and $\mathcal{H}\left[\cdot \right]$ computes the entropy of a probability distribution. Thus, we have the two recognition networks $q_{\phi}(z|x,y)$ and $q_{\phi}(y|x)$. This form of the lower bound has data fit terms for both $x$ and $y$ in the generative model, as well as regularization terms for both recognition networks $q_{\phi}(z|x,y)$ and $q_{\phi}(y|x)$. 

The recognition network $q_{\phi}(z|x,y)$ is shared by both the unlabeled and labeled objectives. The recognition network $q_{\phi}(y|x)$ is unique to the unlabeled data. Following the work in \cite{kingma2014semi,maaloe2016auxiliary}, we add a weighted term to the final objective function to ensure that $q_{\phi}(y|x)$ is trained on all data such that
\begin{align}\label{eqn:objective}
\mathcal{L}(\theta, \phi) & = \sum_{(x,y)\sim \tilde{p}_l}\mathcal{L}^l (x,y;\theta, \phi) + \sum_{x\sim \tilde{p}_u} \mathcal{L}^u(x; \theta, \phi) +\nonumber\\  
& \quad\,\,\alpha \mathbb{E}_{(x,y)\sim \tilde{p}_l} \left[ \log q_{\phi}(y|x)\right]\,,
\end{align}
where $\alpha$ is a small positive constant which is initialized in a similar way as in \cite{kingma2014semi}, $\tilde{p}_l$ is the empirical distribution of labeled points and $\tilde{p}_u$ is the empirical distribution of unlabeled points.

\subsection{Discrete Outputs}

Optimizing Eq. (\ref{eqn:objective}) is straightforward when $y$ is continuous as we can approximate expectations by sampling from $q_{\phi}(z|x,y)$ and $q_{\phi}(y|x)$ and using stochastic backpropagation and the reparameterization trick \cite{kingma2013auto,rezende2014stochastic} to yield differentiable estimators. 

Despite recent efforts \cite{jang2016categorical}, reparameterization for discrete variables is as yet not a well-understood process. In the case where $y$ is a discrete variable, that is, $y \in \{1,...,K\}$, optimization of $\mathcal{L}^u$ requires approximating expectations w.r.t. $q_{\phi}(y|x)$. Rather than using Monte-Carlo approximations for this, we propose to directly compute the expectation by summing over the possible values of $y$:
\begin{align}
\mathbb{E}_{q_{\phi}(y|x)}\left[f(y,z,x) \right] = \sum\limits_{y \in \mathcal{Y}} q_{\phi}(y|x)f(x,y,z)\,.
\end{align}
Substituting this for the relevant terms in Eq. (\ref{eqn:elbo_u2}) yields:
\begin{align}\label{eqn:discreteError}
\mathbb{E}_{q_{\phi}(y|x)} & \left[ \log p_{\theta}(x,y|z)\right] = \nonumber\\
& \sum\limits_{y\in\mathcal{Y}} q_{\phi}(y|x)\mathbb{E}_{q_{\phi}(z|x,y)} \left[\log p_{\theta}(x,y|z)\right]
\end{align}
\noindent and
\begin{align}\label{eqn:discreteKL}
\mathbb{E}_{q_{\phi}(y|x)} &\left[\text{D}_\text{kl}(q_{\phi}(z|x,y)||p(z))\right] =\nonumber \\ 
& \sum\limits_{y\in\mathcal{Y}}q_{\phi}(y|x)\text{D}_\text{KL}\left(q_{\phi}(z|x,y)||p(z)\right)\,.
\end{align}
Taking explicit expectations w.r.t. $q_{\phi}(y|x)$ rather than Monte-Carlo approximations allows us to extend the training procedure to cases where $y$ is discrete. 

\begin{figure*}[!t]
\centering
\begin{subfigure}{.29\textwidth}
  \centering
  \includegraphics[width=.9\linewidth]{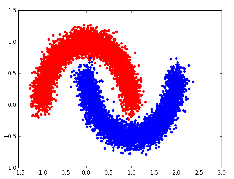}
  \caption{}
  \label{fig:moons}
\end{subfigure}%
\begin{subfigure}{.29\textwidth}
  \centering
  \includegraphics[width=.9\linewidth]{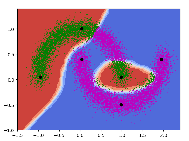}
  \caption{}
  \label{fig:dnn}
\end{subfigure}
\begin{subfigure}{.29\textwidth}
  \centering
  \includegraphics[width=.9\linewidth]{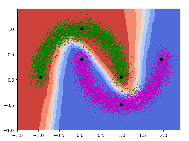}
  \caption{}
  \label{fig:gssl}
\end{subfigure}%
\\
\begin{subfigure}{.29\textwidth}
  \centering
  \includegraphics[width=.9\linewidth]{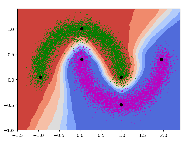}
  \caption{}
  \label{fig:bgssl}
\end{subfigure}%
\begin{subfigure}{.29\textwidth}
  \centering
  \includegraphics[width=.9\linewidth]{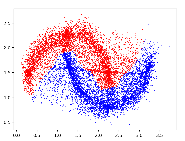}
  \caption{}
  \label{fig:samples}
\end{subfigure}
\caption{Preliminary experiments carried with the model. (a) Complete set of labeled data. (b) Contour plots learned by a standard DNN using only six labeled  labeled data. (c) Contour plots learned by the model using only the depicted points with labels (the rest unsupervised). (d) Contour plots learned with Bayesian training of the model. (e) Samples from the generative model after training.}
\label{fig:fig}
\end{figure*}

\subsection{Introducing Model Uncertainty}

A major advantage of this approach is that it enables us to express model uncertainty in the discriminative component 
$NN_y(z, x, \theta)$ by computing a posterior distribution on the weights $W_y$. For this, we consider the following prior and likelihood functions:
\begin{align}\label{eqn:model_with_w}
p_{\theta}(W_y) & = \mathcal{N}(W_y; 0, \mathbf{I})\,, \\ p_{\theta}(y|x,z,W_y) & = \text{Cat}(y; \pi_y)\,,
\end{align}
where $\pi_y = NN_y(z,x,W_y)$ is parameterized by a neural network with weights $W_y$. Assuming a single labeled data point, the posterior distribution for the latent variables is:
\begin{equation}
p_{\theta}(W_y, z | x, y) = \frac{p_{\theta}(x, y | z,  W_y)p(z)p(W_y)}{p(x,y)}\,.
\end{equation}
This posterior distribution is intractable. Following the work by \citet{blundell2015weight, depeweg2016learning}, we introduce an approximate posterior distribution for $W_y$ given by $q_{\phi}(W_y) = \mathcal{N}(W_y; \mu_w, \sigma^2_w)$ with $\sigma^2_w$ being a diagonal covariance matrix. Note that we are assuming here that $W_y$ is \emph{a posteriori} independent of $z$. We also assume \emph{a posteriori} independence between $W_y$ and $y$ in the unlabeled case.
Re-deriving the lower bound for the case of one single labeled data point yields:
\begin{align}\label{eqn:elbo_bayes_labeled}
\mathcal{L}(x, y; \theta, \phi) & = \mathbb{E}_{q_{\phi}(z,|x,y)q_{\phi}(W_y)}\left[\log p_{\theta}(x,y|z,W_y)\right] - \nonumber\\ 
& \quad \,\, \text{D}_\text{KL}\left(q_{\phi}(z|x,y) \| p(z) \right) - \nonumber \\
& \quad\,\, \text{D}_\text{KL}\left(q_{\phi}(W_y) \| p(W_y) \right)\,.
\end{align}
The corresponding derviation for the unlabeled lower bound can be obtained from
Eq. (\ref{eqn:elbo_u2}) in a similar manner. We follow the work presented by \citet{blundell2015weight}, and optimize the objective functions applying reparameterization to the weights $W_y$ as well as $z$.

\subsection{Prediction with the Model}

To approximate $p(y_\star|x_\star)$
for a new example $x_\star$, 
we have to integrate $p_{\theta}(y_\star|x_\star,z,W_y)$ with respect to the posterior distribution on $z$ and $W_y$. For this, $W_y$ is sampled from $q_{\phi}(W_y)$, while $z$ is sampled from the recognition network
$q_{\phi}(z|x_\star, y_\star)$. Since this recognition network requires $y_\star$, we use a Gibbs sampling procedure, drawing the first sample of $y_\star$ from the recognition network
$q_{\phi}(y_\star|x_\star)$. In particular,
\begin{align}\label{eqn:gibbs}
& \quad y_\star^{(0)}  \sim q_{\phi}(y_\star|x_\star)\,,\nonumber\\ 
&\,\,\left.\begin{array}{r@{\,\,\sim\,\,}l}
W_y^{(\tau)} & q_{\phi}(W_y)\,,\\
z^{(\tau)} &q_{\phi}(z|x_\star, y_\star^{(\tau-1)})\,,\\
y_\star^{(\tau)} & p_{\theta}(y_\star| x_\star, z^{(\tau)},W_y^{(\tau)})\,,
\end{array}\right\} \text{for $\tau=1,\ldots,T$. }\nonumber
\end{align}
with the final prediction being an average over the samples. Using $q_{\phi}(y_\star|x_\star)$ to initialize the procedure increases efficiency and negates the need for a burn in period. In our experiments $T=10$ produced good results.

\section{Preliminary Results}

In this section we detail preliminary results achieved by the proposed model. We experiment with toy data similar to that used by \citet{maaloe2016auxiliary}. The data consists of 1e4 training and test samples, generated from a deterministic function with additive Gaussian noise (Figure \ref{fig:moons}). A small set is selected as the labeled data, and the rest are unlabeled. We compare performance to that of a feed forward neural network.

All neural networks have two hidden layers with 128 neurons, and $z \in \mathbb{R}^{5}$. We set $\alpha = 0.1 * N_l$, where $N_l$ is the number of labeled data points, and use RELU activations for all hidden layers.  

The model converges on 100\% accuracy in all cases for $N_l > 10$. When examining training curves for different values of $N_l$ (not shown due to space constraints), we see that as the labeled set is larger, training converges faster and to better lower bounds.

\begin{table}[t]
\caption{Test accuracy and log-likelihood for each method.}
\label{sample-table}
\vskip 0.15in
\begin{center}
\begin{small}
\begin{sc}
\begin{tabular}{lccc}
\hline
 & DNN & SSLPE & SSLAPD\\
\hline
Log-likelihood  & -1.20 & -0.07 & -0.01 \\
Accuracy        & 83.6  & 99.2  & 99.7  \\
\hline
\end{tabular}
\end{sc}
\end{small}
\end{center}
\vskip -0.1in
\end{table}

Figures \ref{fig:dnn}, \ref{fig:gssl}, \ref{fig:bgssl} show, respectively, the predictive probabilities learned by a deep neural network (DNN) which ignores the unlabeled data, the proposed approach for Semi-Supervised Learning using a Point Estimate for $W_y$ (SSLPE) and the proposed approach for Semi-Supervised Learning using an Approximate Posterior Distribution for $W_y$ (SSLAPD).
We use $N_l=6$ in all cases: three labeled examples selected from each class, in a similar manner to \citet{maaloe2016auxiliary}. Table \ref{sample-table} shows the average test log-likelihood and predictive accuracy obtained by each method.
DNN overfits the labeled data and achieves low predictive accuracy and test log-likelihood. SSLPE is able to leverage the unlabeled data to learn a smoother decision boundary that aligns with the data distribution, achieving
much better predictive performance than DNN. Finally, SSLAPD makes predictions similar to those of SSLPE but with higher uncertainty in regions far away from the training data. Overall, SSLAPD is the best performing method.
Finally, Figure \ref{fig:samples} shows samples generated by SSLAPD, indicating this method has learned a good generative process and is able to synthesize compelling examples.

\section{Discussion and Future Work}

DGMs have been successfully applied to semi-supervised learning tasks, though by discarding the model and using only an inference network for predicting labels. This approach does not allow to account for model uncertainty.

In contrast, the proposed approach uses a Bayesian neural network for label prediction. In addition to being more satisfying from a modeling perspective, this opens the door to joint semi-supervised and active learning by accounting for model uncertainty.

Our experiments show that the proposed approach is promissing and able to produce wider confidence bands far away from the training data than alternative methods that ignore parameter uncertainty.
Further experiments with alternative datasets such as MNIST are required. Future work includes developing  methods for acquiring new samples from a pool set, performing joint semisupervised and active learning, and comparing with recent benchmark methods.

\bibliography{DGML}
\bibliographystyle{icml2017}

\end{document}

%% file: vae.tex
\begin{tikzpicture}
  \node[obs] (x) {$x$};%
  \node[latent,above=of x] (z) {$z$}; %
  \plate [inner sep=.3cm,xshift=.02cm,yshift=.2cm] {plate1} {(x) (z)} {$N$}; %
  \edge {z} {x} 
\end{tikzpicture}

%% file: cvae.tex
\begin{tikzpicture}
\node[latent] (y) {$y$};%
\node[latent,left=of y] (z) {$z$}; %
\node[obs,below=of y] (x) {$x$}; %
\plate [inner sep=.3cm,xshift=.02cm,yshift=.2cm] {plate1} {(x) (z) (y)} {$N$}; %
\edge {z, y} {x}
\end{tikzpicture}

%% file: stochastic_bnn.tex
\begin{tikzpicture}
\node[obs] (x) {$x$};%
\node[latent,left=of x] (z) {$z$}; %
\node[latent,below=of x] (y) {$y$}; %
\plate [inner sep=.3cm,xshift=.02cm,yshift=.2cm] {plate1} {(x) (z) (y)} {$N$}; %
\edge {z, x} {y}
\end{tikzpicture}